\documentclass[runningheads]{llncs}
\usepackage{graphicx}
\usepackage{cite}
\usepackage{amsmath,amssymb,amsfonts}
\usepackage{algorithm}
\usepackage{algorithmic}
\usepackage{subfigure}
\usepackage{textcomp}
\usepackage{xcolor}
\usepackage{booktabs}
\usepackage{enumerate}
\usepackage{url}
\begin{document}
\title{Anomaly Subsequence Detection with Dynamic Local Density for Time Series}
%
%
\author{Chunkai Zhang\inst{1} \and
	Yingyang Chen\inst{2} \and
	Ao Yin\inst{3}
}
\authorrunning{Yingyang Chen et al.}
%
\institute{ Department of Computer Science and Technology, \\
	Harbin Institute of Technology, Shenzhen, China\inst{1,2,3} \\
	\email{ckzhang812@gmail.com\inst{1}, yingyang\_chen@163.com\inst{2}, 
		yinaoyn@126.com\inst{3}
		}
}
\maketitle              
\begin{abstract}
Anomaly subsequence detection is to detect inconsistent data, which always contains important information, among time series. Due to the high dimensionality of the time series, traditional anomaly detection often requires a large time overhead; furthermore, even if the dimensionality reduction techniques can improve the efficiency, they will lose some information and suffer from time drift and parameter tuning.	
In this paper, we propose a new anomaly subsequence detection with Dynamic Local Density Estimation (DLDE) to improve the detection effect without losing the trend information by dynamically dividing the time series using Time Split Tree. In order to avoid the impact of the hash function and the randomness of dynamic time segments, ensemble learning is used. Experimental results on different types of data sets verify that the proposed model outperforms the state-of-art methods, and the accuracy has big improvement.

\end{abstract}

\section{Introduction}
\label{sec:intro}
The time series data is stored in the order of the data generation time, and is dynamic and massive. We are interested in finding the abnormal subsequence in complete time series,
in other words, anomaly subsequences are inconsistent with the shape of most other subsequences. 
Anomaly detection for time series is an analysis of inconsistent data with normal data, which always represents an emergency or fault. Itc is applied in many application domains, ranging from financial data\cite{Ruiz2012Correlating,Rahmani2014Graph}, Electrocardiogram (ECG) data\cite{Argyro2009Heartbeat,Sivaraks2015Robust} to sensor data\cite{Lazaridis2003Capturing}. For example, analysis of ECG data can timely monitor patients' heart health such as arrhythmia, ventricular atrial hypertrophy, myocardial infarction\cite{Ocak2009Automatic} before diagnosis
process. Therefore, timely detection of abnormal data contained in the data is of great significance.

A rich body of literature exist on detecting time series anomalies, however, existing anomaly detection methods\cite{Ren2018Anomaly,Sun2014An,Lippi2013Short,Ren2017A} still suffer from a lot of problems. 
Time series is often high-dimensional data, therefore the calculations in the original data storage format often require large storage and computational overhead. 
In recent years, the different time series data representation methods were proposed to achieve the purpose of dimensionality reduction. Discrete Fourier Transformation (DFT)\cite{faloutsos1994fast} can convert time series of length $n$ into $m$ coefficients by discrete Fourier transform method; Discrete Wavelets Transformation (DWT)\cite{Chan1999Efficient} is a multi-resolution representation of the data signal but can only be used in time series of integer powers of length $2$; and Piecewise Aggregate Approximation (PAA)\cite{Keogh2001Dimensionality} divides the time series into equal length segments, then takes the average for each segment. As for Symbolic Aggregate Approximation (SAX)\cite{Lin2003A}, it maps the mean of the segments to a symbolic representation based on PAA as other variants, ESAX\cite{Lippi2013Short} and SAX-TD\cite{Sun2014An}. 
All these methods can reduce the dimensionality but losing information on local time segment.
However, there are some problem that the size of the window needs to be set manually, which requires the relevant expert knowledge \cite{Sivaraks2015Robust}. And the average in the sliding window will lose some important information.
In addition, these methods have not pay much attention to time drift problem, which will get wrong anomaly subsequence if using Euclidean distance, and the details will be discuss in Section \ref{sec:problem}.

We also need to perform anomaly calculations on the representation of time series. The simplest and straightforward method of anomaly subsequence detection is to calculate the similarity between each pair of subsequences by double-loop violence, and treat the most dissimilar subsequences with most other subsequences as abnormal subsequences\cite{keogh2005hot}. In order to improve the efficiency of the brute force algorithm, Keogh et al. proposed HOT SAX\cite{keogh2005hot} to construct an index tree using SAX symbol sequences to optimize the search order of candidate. Li et al.\cite{li2013finding} proposed BitClusterDiscord, who used binary representation to approximate the trend information then use K-media clustering and two pruning strategies to reduce the number of similarity calculations. Senin et al.\cite{senin2015time} proposed Rare Rule Anomaly to discrete the time series into symbol and derive context-free grammar to discover algorithmic irregularities associated with exceptions. Ren et al. proposed PAPR-RW\cite{Ren2017A} based on PAPR representation and random walk model\cite{Moonesinghe2006Outlier} to convert time series into similar matrices. 
All these method use sliding window to split time series into subsequence while set the size of window manually. Once the window setting is not good enough to different kind of data sets, it is easy to detect wrong anomaly subsequence. 

In this paper, we propose a novel anomaly subsequence detection of Dynamic Local Density Estimation (DLDE) where TSTree is used to dynamically divide the time series, and hash function to improve the efficiency. In order to avoid the influence of the hash function and the randomness of dynamic time segments, ensemble learning is used in our method. And this algorithm can improve the effect of detection without losing the time series trend information by dynamic segment and has less parameters.

The contribution of this paper can be summarized as follows.
\begin{itemize}
	\item[(1)] An anomaly detection algorithm is proposed to solve the time drift problem inspired by  the idea of DTW. And the detection effect can be improved without losing the trend information because this algorithm does not compress the original time series.
	\item[(2)] We propose a novel data structure named Time Split Tree (TSTree) and introduce the three techniques in DLDE, Time Split Tree for time series randomly division, Hash Table for similarity measurement that the data points with the same hash value are similar data points, and Ensemble Learning to ensure the stability of algorithm. 
	\item[(3)] Our algorithm is analyzed with solid theoretical explanation and experimentally verified the effectiveness of the algorithm. DLDE outperforms other state-of-art algorithms on different types of data sets in accuracy. 
\end{itemize}

The rest of paper is organized as follows. Section \ref{sec:problem} sets up the problem definitions for anomaly detection in time series. Section \ref{sec:algorithm} proposes the Dynamic Local Density Estimation algorithm. Experimental results are reported in Section \ref{sec:experiment}. Finally, Section \ref{sec:conclusion} concludes the paper.

\section{Problem Statement}\label{sec:problem}
\begin{figure}[t] \centering 
	\subfigure[] { \label{fig:dtwr}
		\includegraphics[width=0.3\textwidth]{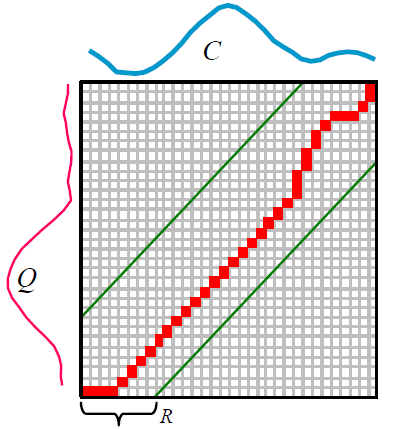} 
	} 
	\subfigure[] { \label{fig:dtw}
		\includegraphics[width=0.4\textwidth]{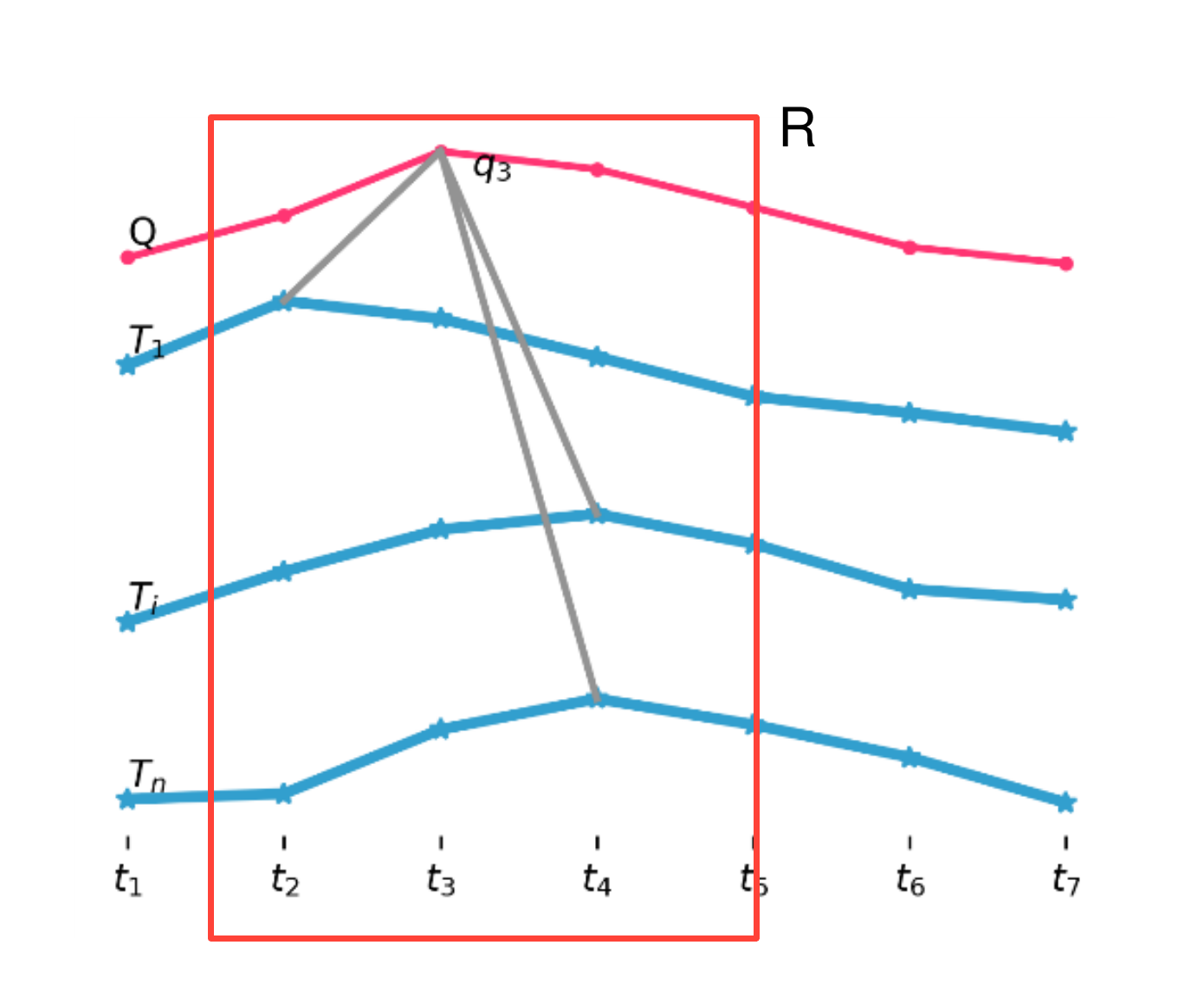} 
	} 
	
	\caption{The Fig. (a) is the DTW calculation matrix with adjustment window $R$(green window), $C$ and $Q$ are the two time series. The Fig. (b) is the example of DTW calculates schematics in all data sets.} 
	\label{fig:example}
\end{figure}

Dynamic time warping (DTW) is a dynamic programming technique which can handle nonlinear alignments and local drift time \cite{dtwdrifttime} with different length subsequences caused by timeline scaling, amplitude shift and linear drift. 
Amplitude shift is ampliotude baseline is different with two similar time series. Timeline scaling means time series scaling proportionally on the timeline. Linear drift shows a trend od linear increasing or decreasing for time series.
If the corresponding subsequences in two time series do not represent the same meaning, it is unreasonable to calculate their similarity by means of Euclidean distance.
In order to reduce the time complexity, warping function\cite{Rakthanmanon2012Searching} was proposed as shown in Fig.\ref{fig:dtwr}. After adding the optimization width limit, the most similar data points can be found only within a certain segment.


The anomaly detection based on DTW needs to calculate the similarity of any two subsequences and the time complexity is $O(mnN^2)$. If adding the search scope limit window R, the calculation process is shown in Fig.\ref{fig:dtwr}. 
In Fig.\ref{fig:dtw}, suppose we should detect whether the time series $Q$ has anomaly or not, and the other series are $T_1 \sim T_n$, the adjustment window is $R$. Take the $q_5$ as an example, finding the minimum distance in the limited R window from $T_1$ to $T_n$. 
From the perspective of anomaly detection, the larger of distance between $q_5$ and other data points, the more abnormal the point is. In other words, if there is no similar data point in the adjustment window, the test point should be an anomaly. 
Therefore, inspired by the idea above, we propose a method to quickly evaluate the similarity of subsequence. Based on this method, an anomaly subsequence detection algorithm for dynamic local density estimation is proposed.

%

\section{The Proposed Algorithm} \label{sec:algorithm}
Based on the analysis of dynamic time warping similarity calculation in Section \ref{sec:problem}, we propose a time series anomaly subsequence detection algorithm, Dynamic Local Density Estimation(DLDE)
to divide the time series randomly and evaluate the degree of anomaly for data points through dynamic local density of each data point in the subsequence.

\subsection{Basic Concept and Definitions}\label{sec:def}

\begin{definition}[Time Split Tree(TSTree)]
	TSTree randomly divides a time series into several dynamic time segments, each of which is located at the leaf node. 
\end{definition}
The process is as follows: there is a time series $\{t_1 \sim t_d\}$, randomly choose time point $st$ as a split point, and divide all time points before $st$ into $T_l$ while others in $T_r$. Recursively the above process until the stop conditions:
\begin{itemize}
	\item[1)] The length of the time segment at the leaf node is less than or equal to 3.
	\item[2)] The depth of the tree is equal to $log_2(d)$.
\end{itemize}

\begin{figure}[t]
	\centering
	\includegraphics[width=0.4\linewidth]{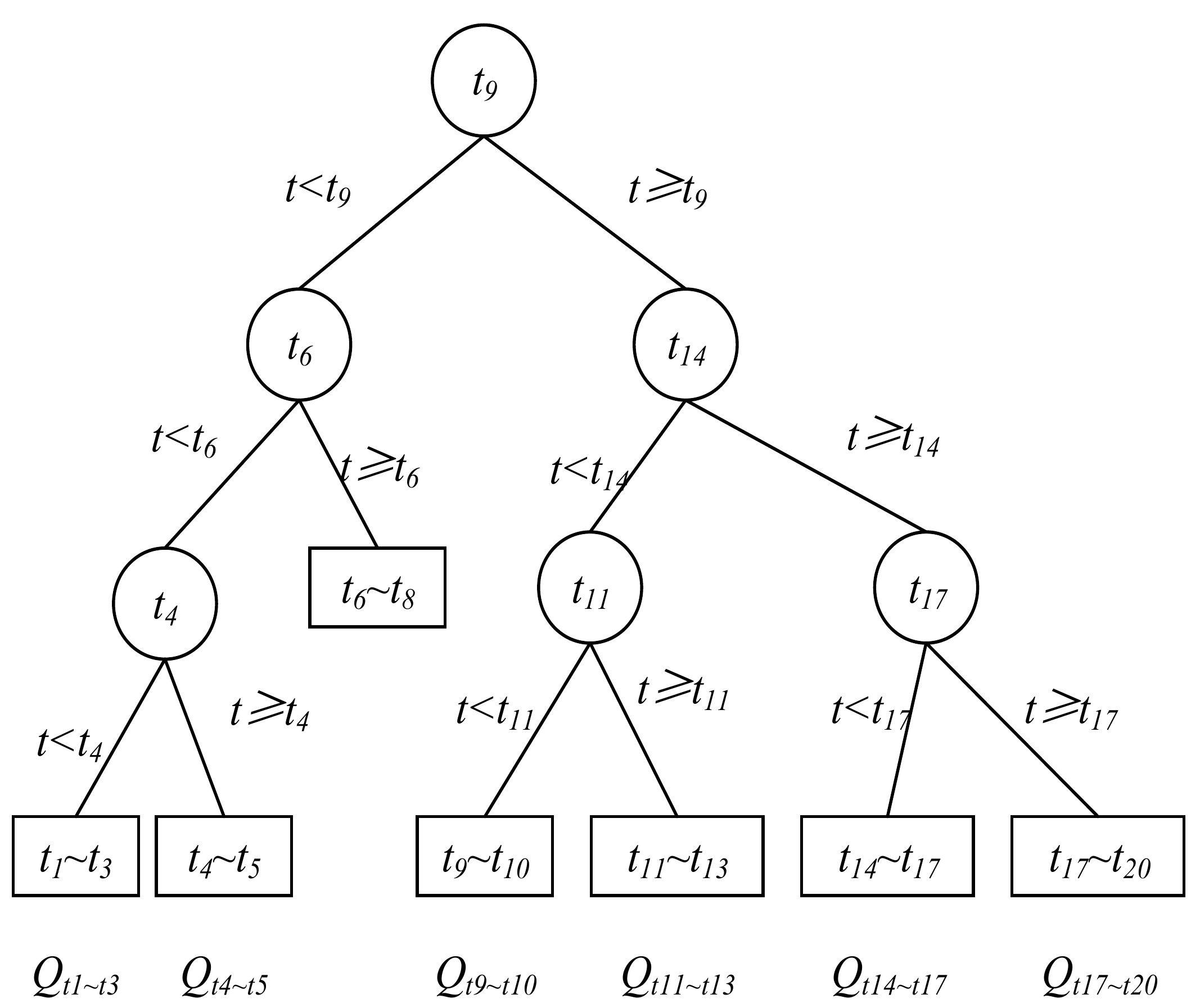}
	\caption{The structure of TSTree, the circle node represents an internal node, and a rectangle node represents a leaf node.}
	\label{fig:tree}
\end{figure}

Give an example of the TSTree. Assuming that the time points of the Q time series are $t_1$ to $t_{20}$, and the divided result is shown in Fig.\ref{fig:tree}. Select $t_9$ as the split node for root, and divide $t_1 \sim t_8$ to left subtree, and $t_9 \sim t_{20}$ to the right subtree.

\begin{definition}[Dynamic Time Segment R]
	Inspired by limit window R in DTW in Fig.\ref{fig:example}, Dynamic time segment refers to a continuous time segment in a subsequence that is used to find the most similar data points, such as $R=\{t_s,t_{s+1},...,t_e\} (1 \leq s \leq e)$. 
	
\end{definition}

\begin{definition}[Hash Function] \label{hash function}
	The data set $Q_{t_1 \sim t_d}$ at $d$ time points can be mapped to $d$ hash table $HashTable_{t_1 \sim t_d}$ by hash function(Equation (\ref{eq:hash})). If two data points have the same hash function value, the two data points are similar. 
	\begin{equation} \label{eq:hash}
	hash(p)=\lfloor \frac{p+r}{w} \rfloor 
	\end{equation}
	where $p$ is the time point, $w$ is the hash function width parameter randomly sampled from the range $[1.0/log_2(N),1-1.0/log_2(N)]$, and $r$ is a parameter randomly selected from the range $[0,w]$.
\end{definition}

\begin{definition}[Similarity Time Point Set] \label{simtimepointset}
	Suppose $p_r$ is the value of time point $t_r$ and there is a dynamic time segment $R=t_s,...,t_e$ and the corresponding dataset $Q_{t_s \sim t_e}$ with $HashTable_{t_s \sim t_e}$. The similarity time point set is calculated as
	\begin{equation} \label{eq:sim}
	N(p_r) = \{t_j|t_j\in[t_s,t_e],hash(p_r)\in HashTable_{t_j}\}
	\end{equation}
\end{definition}

\begin{definition}[True Similarity Relation]\label{truesimrela}
 Due to the randomness of the hash function, the set $N(p_r)$ may contains points that are not true similarity relationship with $p_r$. 
	Therefore, $h$ random hash function are used to find the intersection of $N(p_r)$, which is the true similarity relation set as shown in Equation (\ref{eq:real}).
	\begin{equation} \label{eq:real}
	TN(p_r)= N_1(p_r) \cap N_2(p_r) \cap ... \cap N_h(p_r)
	\end{equation}
\end{definition}

\begin{definition}[Local Density]\label{localdensity}
	Local density $density(Q_{t_j},q_i)$ refers to the number of similar data points $q_i$ in the data point set $Q_{t_j}$.
	\begin{equation}
	\begin{split}
	density(Q_{t_j},q_i)=count\{hash(Q_{k,t_j})=hash(q_i) | k<N, Q_{k,t_j}\in Q_{t_j}\}
	\end{split}
	\end{equation}
\end{definition}

\begin{definition}[Dynamic Local Density]\label{dynamiclocaldensity}
	Dynamic Local Density refers to evaluating the local density of data points $q_i$ in corresponding dynamic time window 
	\begin{equation}
	Density(q_i)=\frac{1}{|TN(q_i)|}\sum_{t_j \in TN(q_i)}density(Q_{t_j},q_i)
	\end{equation}
\end{definition}

\subsection{Anomaly Detection Algorithm in Time Series} \label{sec:dlde}
The above section introduces the proposed definition and data structure, in this section, we are going to introduce the dynamic local density estimation, which is the core of the our algorithm. To determine the anomaly of the time series, we evaluate the local density of time series by evaluating the local density of each data point within the dynamic time segment. 


\textbf{1) Divide dynamic time segment.}

Dynamic density estimation is to evaluate the local density of data points through dynamic time segments. Therefore, dividing the time series into multiple disjoint time segments is the first step. We randomly construct TSTrees to dynamically divide time series and each leaf contains one segment. The pseudo-code are shown in Algorithm \ref{alg:A}.
\begin{algorithm}[h]
	\caption{Build TSTree (Init\_TSTree)}
	\label{alg:A}
	\begin{algorithmic}[1]
		\REQUIRE ~~\\ 
		Time Series Data Set $Q_{t_1-t_d}$;\\
		First Time Point, $t_1$, The End Time Point, $t_d$;\\
		Hight Limit, $hlimit$, Size Limit, $slimit$; \\
		Current tree height, $height_{cur}$;
		\ENSURE ~~\\ 
		A Time Split Tree, $TSTree$;
		\IF {$t_d$-$t_1 \leq slimit$ or $height_{cur} \geq hlimit$}
		\STATE {Return TreeNode($t_1\texttildelow t_d$);}
		\ENDIF
		\STATE {Randomly select a split time point, $st$}
		\STATE {Build Left Tree, Init\_TSTree($t_1$,  $st-1$, $hlimit$, $slimit$)}
		\STATE {Build Right Tree, Init\_TSTree($st$,  $t_d$, $hlimit$, $slimit$)}
		\STATE {Return $TSTree$;} 
	\end{algorithmic}
\end{algorithm}

\textbf{2) Build a hash table.}

After dividing the time series into dynamic time segments, we need to use hash function to map data points to hash table in each segment, which can quickly estimate the local density of data points. Suppose the time segment on a leaf node in TSTree is $t_s \sim t_e$.
First, $h$ number of hash functions should be generate as ${\{hash_1(.),hash_2(.),...,hash_h(.)\}}$ following the Equation (\ref{eq:hash}). Then, all these hash functions can map leafs to $h$ number of hash tables.
Each hash table is a two-dimensional array as Equation (\ref{eq:hashtable2}), and each element in the hash table is stored in the form of Key-Value, \textit{Key} $(key_{i,r})$ represents the hash value, and \textit{Value} $(val_{i,r})$ represents the number of times this hash value appears in the data set. The bigger of $val_{1,s}$, the more data points will be map to $key_{1,s}$ at $t_s$, and the more likely the corresponding original data point is normal; otherwise, the smaller of $val_{1,s}$, the more likely the original data is anomaly.
The width is equal to the length of the time segment contained in the leaf node, and the length of each column may be different.
\begin{equation} \label{eq:hashtable2}
HashTable_j=
\left[
\begin{array}{ccccc}
(key_{1,s},val_{1,s}) & ... & (key_{1,r},val_{1,r}) &...&(key_{1,e},val_{1,e})\\
... &  & ... & &...\\
(key_{k,s},val_{k,s}) & ... & (key_{k,r},val_{k,r}) &...&(key_{k,e},val_{k,e})\\
... &  & ... & &...\\
(key_{{x_1},s},val_{{x_1},s}) & ... & (key_{{x_r},r},val_{{x_r},r}) &...&(key_{{x_e},e},val_{{x_e},e})\\
\end{array} 
\right ]
\end{equation}

%
%

The above process uses one hash function to map one leaf node data. In order to calculate the true similarity of the data points on the leaf nodes, $h$ hash tables need to be constructed for each node. 

\textbf{3) Calculate the dynamic local density of data points}

The formula for calculating the dynamic local density of a data point is described in Definitions \ref{dynamiclocaldensity}. And the Algorithm \ref{alg:B} describes the detailed calculation process after a dynamic time segmentation. 

\begin{algorithm}[h]
	\caption{Calculate the local density at each time point in the time series.}
	\label{alg:B}
	\begin{algorithmic}[1]
		\REQUIRE ~~\\ 
		Time Point, $p_i$,corresponding time $t_i$ and TSTree, $tree$;\\
		\ENSURE ~~\\ 
		The local density of $p_i$, $Density(p_i)$;
		\STATE $Density(p_i)=0$;
		\STATE Query the leaf node where $t_i$ located;
		\STATE Leaf node $t_s$ contains the start time point of the time period;
		\STATE Leaf node $t_e$ contains the end time point of the time period;
		\STATE $Hash(.)=\{hash_1(.), hash_2(.), ..., hash_h(.)\}$;// H hash functions are contained in;
		\STATE {$TN(q_i) \leftarrow $ a collection of all time points from $t_s$ to $t_e$;}
		\FOR {each $hash_j(.)$ in $Hash(.)$}
		\STATE {$ksy_{i,j}=hash_j(q_i)$;// calculate the hash value of $p_i$;}
		\FOR {each $t=t_s$ to $t_e$}
		\IF {$k_{i,j}$ in $HashTable_{i,t}(.) \rightarrow ksys()$}
		\STATE {$N_j(q_i) \leftarrow t$;}
		\ENDIF
		\ENDFOR
		\STATE {$TN{q_i} \leftarrow TN(q_i) \cap N_j(q_i)$;}
		\ENDFOR
		\FOR {$t$ in $TN(q_i)$}
		\FOR {each $hash_j(.)$ in Hash(.)}
		\STATE {$key_{i,j}=hash_j(q_i)$;}
		\STATE {$Density(p_i)+=HashTable_{j,t} \rightarrow get(key_{i,j})$}
		\ENDFOR
		\ENDFOR
		\STATE {Return $Density(p_i)$;} 

	\end{algorithmic}
\end{algorithm}

\textbf{4) Calculate the local density of the subsequence}

Step 3 completes the dynamic local density estimation of a data point; then the local density of the time series $P$ is estimated as shown in Equation (\ref{eq:density}), where $d$ is the length of time series $P$ and $Density(p_i)$ is calculated by Definition \ref{dynamiclocaldensity}. We can see that if the $Density(P)$ value is larger, it indicates that the data points in the time series $P$ are similar to most of the time series data points in the data set, therefore, the time sequence $P$ is more likely to be a normal time series.
\begin{equation} \label{eq:density}
Density(P)=\frac{1}{d}\sum_{i=1}^{d}Density(p_i)
\end{equation}

\textbf{5) Use Ensemble learning to determine the anomaly}

Steps 1 to 4 evaluate the anomaly of each subsequence in the data set by dividing the subsequences into disjoint dynamic time segments once. However, since the data stored by TSTree is randomly segmented, if there is only one TSTree, the algorithm will not get a stable calculation result. Therefore, the idea of using ensemble learning is proposed to construct $m$ TSTrees to form \textbf{TSForest}. The score of the subsequence $P$ is calculated by TSForest as the dynamic density mean of $m$ TSTree evaluations, and the formula for calculating the score of the subsequence is as shown in Equation (\ref{eq:tsforest}). The smaller the subsequence $P$ is, the more likely subsequence $P$ is an abnormal subsequence.
\begin{equation} \label{eq:tsforest}
Score(P)=\frac{1}{m}\sum{Density(P)}
\end{equation}

\begin{algorithm}[h]
	\caption{DLDE anomaly detection algorithm in time series.}
	\label{alg:c}
	\begin{algorithmic}[1]
		\REQUIRE ~~\\ 
		Time Series $P$, Subsequence Length, $s$, Hash Table Number $h$, TSTree Number $m$;\\
		\ENSURE ~~\\ 
		The anomaly score of each subsequence, $Score$;
		
		\STATE {$n \leftarrow $ The length of $P$;}
		\STATE {Dividing the time series $P$ into a time series set $Q$ according to the subsequence length $s$;}
		\FOR {$i=1$ to $m$}
		\STATE {Build TSTree;}
		\STATE {Initialize $h$ hash functions for each leaf node of TSTree;}
		\STATE {Constructing a hash table on each leaf node;}
		\ENDFOR
		\FOR {each subsequence in $Q$}
		\FOR {each TSTree in TSForest}
		\STATE {Calculating $Density(Q_i)$;}
		\ENDFOR
		\STATE {$Score \leftarrow Mean(Density(Q_i))$}
		\ENDFOR
		\STATE {Return $Score$;} 
	\end{algorithmic}
\end{algorithm}

\subsection{Analysis}

\textbf{Time Complexity.}
Suppose the size of time series data set is $N$, and the length of subsequence is $d$. The time to build $m$ TSTree needs $O(m*log_2(d))$ and the time complexity of $h$ Hash Table is $O(N*m*d*h)$, therefore, in the detection process, the time complexity is $O(N*m*d*h*log_2(d))$. It is verified in Section\ref{sec:experiment} that $m$ and $h$ can achieve convergence by taking a small constant algorithm.

\textbf{Space Complexity.}
DLDE takes advantage of the data structure of the TSTress and the Hash Table. The TSForest composed by $m$ TSTrees and the data in every leaf node needs $h$ Hash Tables to represent. Therefore, the space complexity required by the algorithm is $O(m*h*d*const)$, where $const$ represents the number of hash values.

\section{Experimental Evaluation} \label{sec:experiment}
In this section, the data sets and the evaluation metrics are introduced first. For comparability, we implemented all experiments on our workstation with 2.5GHz, 64 bits operation system, 4 cores CPU and 16GB RAM.

\subsection{Evaluation Metrics and Experimental Setup}


\textbf{Data sets:} 
The time series data sets in the experiments are selected from the UCR Time Series Repository\cite{UCRArchive} and the BIDMC Congestive Heart Failure Database\cite{Baim1986Survival}.
In UCR, the ECG data and the SENSOR data set are typical time series data sets; MOTION is the sequence data generated by the action, the IMAGE data can extract the time series data.
These data sets are described in Table \ref{table:ucr}. 
In our experiments, we follow the split subsequences as provided 
by UCR. 
For balanced data, 
we will significantly under-sampling one of two classes to obtain 
minority(anomaly class). For example, in ECG5000\_2\_3 we choose class 2 as normal and class 3 as anomaly.

\begin{table*}[h]
	\centering
	\caption{The description of UCR time series data sets.}
	\label{table:ucr}
	\begin{tabular}{cccccc}
		\hline
		\multicolumn{1}{c}{No.} & \multicolumn{1}{c}{data sets} 
		& \multicolumn{1}{c}{size} & \multicolumn{1}{c}{length} 
		& \multicolumn{1}{c}{anomaly rate} & type \\
		\hline
		1     & DistalPhalanxOutlineCorrect & 876   & 80    & 38.47\% & Image \\
		2     & \multicolumn{1}{c}{ECG200} & 200   & 96    & 33.50\% & 
		ECG \\
		3     & HandOutlines & 1370  & 2709  & 36.13\% & Image \\
		4     & Lighting2 & 121   & 637   & 39.66\% & Sensor \\
		5     & MoteStrain & 1272  & 84    & 46.14\% & Sensor \\
		6     & SonyAIBORobotSurfaceII & 980   & 65    & 38.36\% & Sensor \\
		7     & ToeSegmentation2 & 166   & 343   & 25.30\% & Motion \\
		8     & ECG5000\_2\_3 & 1863  & 140   & 5.15\% & ECG \\
		9     & ECG5000\_2\_4 & 1961  & 140   & 9.89\% & ECG \\
		10    & ECG5000\_2\_5 & 1791  & 140   & 1.34\% & ECG \\
		11    & StarLightCurves\_2\_1 & 427   & 1024  & 35.59\% & Sensor \\
		12    & DiatomSizeReduction\_2\_1 & 132   & 345   & 25.75\% & Image \\
		13    & DiatomSizeReduction\_3\_1 & 133   & 345   & 25.56\% & Image \\
		14    & DiatomSizeReduction\_4\_1 & 125   & 345   & 27.20\% & Image \\
		\hline
	\end{tabular}%
	
\end{table*}%

\textbf{Experimental Setup:} We select five anomaly detection algorithms, 
Relative Density Outlier Score(RDOS)\cite{tang2017local}, Fast Variance Oulier Angle(FastVOA)\cite{pham2012near}, Internal\cite{Ren2018Anomaly} and Piecewise Aggregate Pattern Representation(PAPR)\cite{Ren2017A}. RDOS is the anomaly detection algorithms based on local density, FastVOA is an algorithm based on angle variance. Internal and PAPR are two anomaly detection algorithms based on interval division. The parameter settings of the above comparison algorithm are set according to the reference. For RDOS, the neighbors number will be set to $10$. For FastVOA, we will set the hash number to $100$. For PAPR, we will set the three parameters $wc=0.3, wd=0.4, wr=0.3$. All these compared algorithms and DLDE are executed for 50 times to get stable results. 

\subsection{Accuracy}
The aim of this experiment is to compare DLDE with other methods in terms of Area Under Curve(AUC). AUC is commonly used for evaluating anomaly detection algorithm. The experiment results are recorded in Table \ref{table:auc}, and the best results are highlighted in bold font. $NA$ indicates that this algorithm cannot be calculated on this data set in the current experimental environment. From this table, we can find that DLDE has better results than other algorithms on the most of all data sets(12/14). It is indicated that DLDE is able to detect anomalies efficiently that other baselines are difficult to detect. 

\begin{table}[htbp]
	\centering
	\caption{AUC Performance. The best AUC scores are highlighted in bold.}
	\label{table:auc}
	\begin{tabular}{cccccc}
		\hline
		\multicolumn{1}{c}{No.} & \multicolumn{1}{c}{DLDE} &\multicolumn{1}{c}{RDOS} & \multicolumn{1}{c}{FastVOA} & \multicolumn{1}{c}{Internal} & \multicolumn{1}{c}{PAPR-RW} \\
		\hline
		1     & \textbf{0.705} & 0.646 & 0.702 & 0.632 & 0.693 \\
		2     & \textbf{0.875} & 0.658 & 0.84  & 0.609 & 0.788 \\
		3     & \textbf{0.815} & NA & 0.772 & 0.576 & 0.734 \\
		4     & \textbf{0.764} & 0.611 & 0.697 & 0.599 & 0.653 \\
		5     & \textbf{0.848} & 0.529 & 0.730  & 0.579 & 0.724 \\
		6     & \textbf{0.796} & 0.52  & 0.715 & 0.486 & 0.571 \\
		7     & 0.736 & 0.717 & 0.720  & 0.671 & \textbf{0.777} \\
		8     & \textbf{0.861} & 0.669 & \textbf{0.861} & 0.659 & 0.837 \\
		9     & \textbf{0.735} & 0.591 & 0.716 & 0.668 & 0.714 \\
		10    & 0.870 & 0.904 & \textbf{0.93} & 0.823 & 0.838 \\
		11    & \textbf{0.966} & NA & 0.833 & 0.769 & 0.750 \\
		12    & \textbf{1.00} & 0.652 & 0.970  & \textbf{1.00}     & 0.998 \\
		13    & \textbf{0.901} & 0.774 & 0.853 & 0.677 & 0.761 \\
		14    & \textbf{1.00} & 0.768 & 0.971 & \textbf{1.00} & \textbf{1.00} \\
		\hline
	\end{tabular}%
\end{table}%

\begin{figure}[h]
	\centering
	\subfigure[Comparison results of each algorithm on different length time series data.]{
		\includegraphics[width=0.435\linewidth]{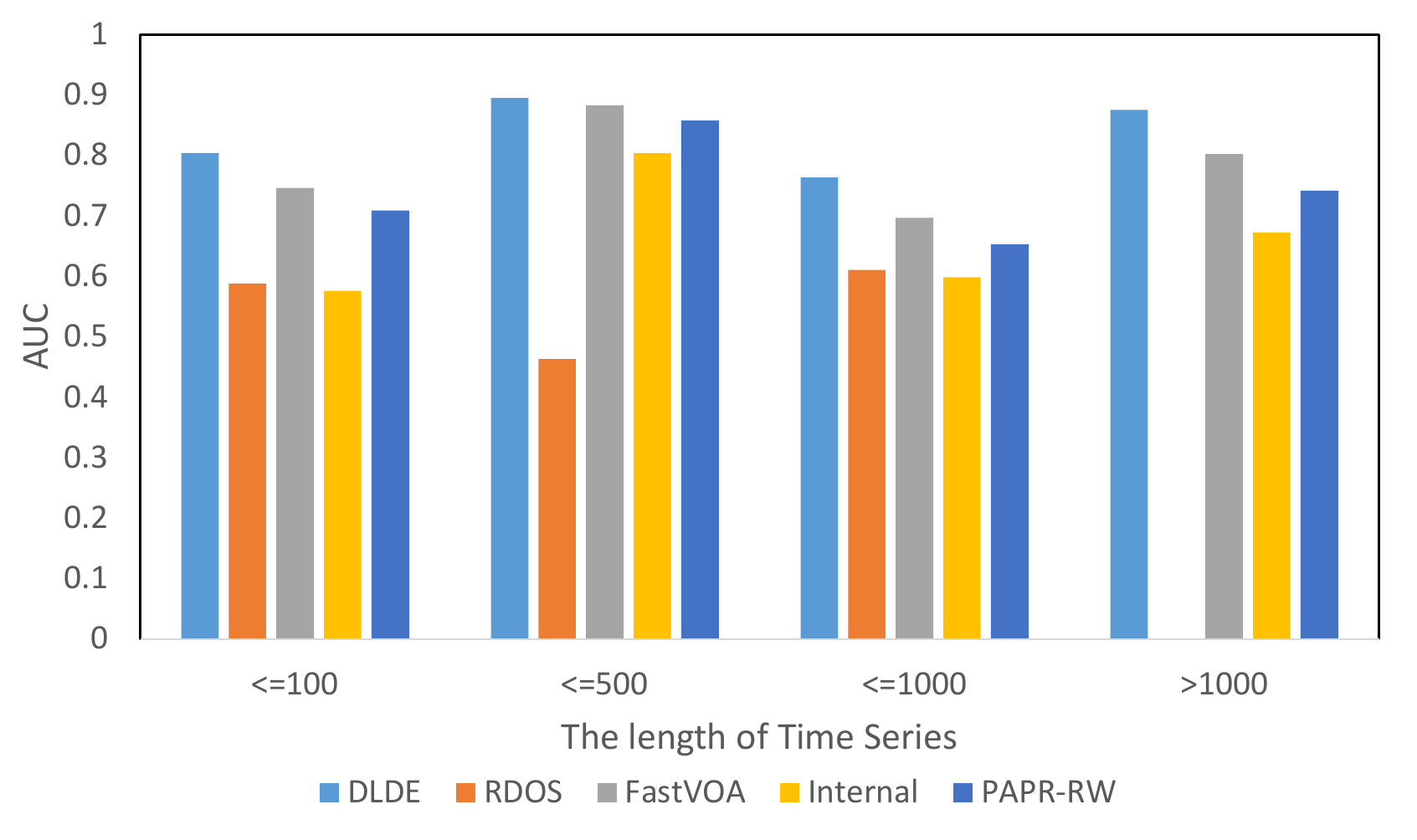}
	}
	\quad
	\subfigure[Comparison results of each algorithm on different type of time series data.]{
		\includegraphics[width=0.45\linewidth]{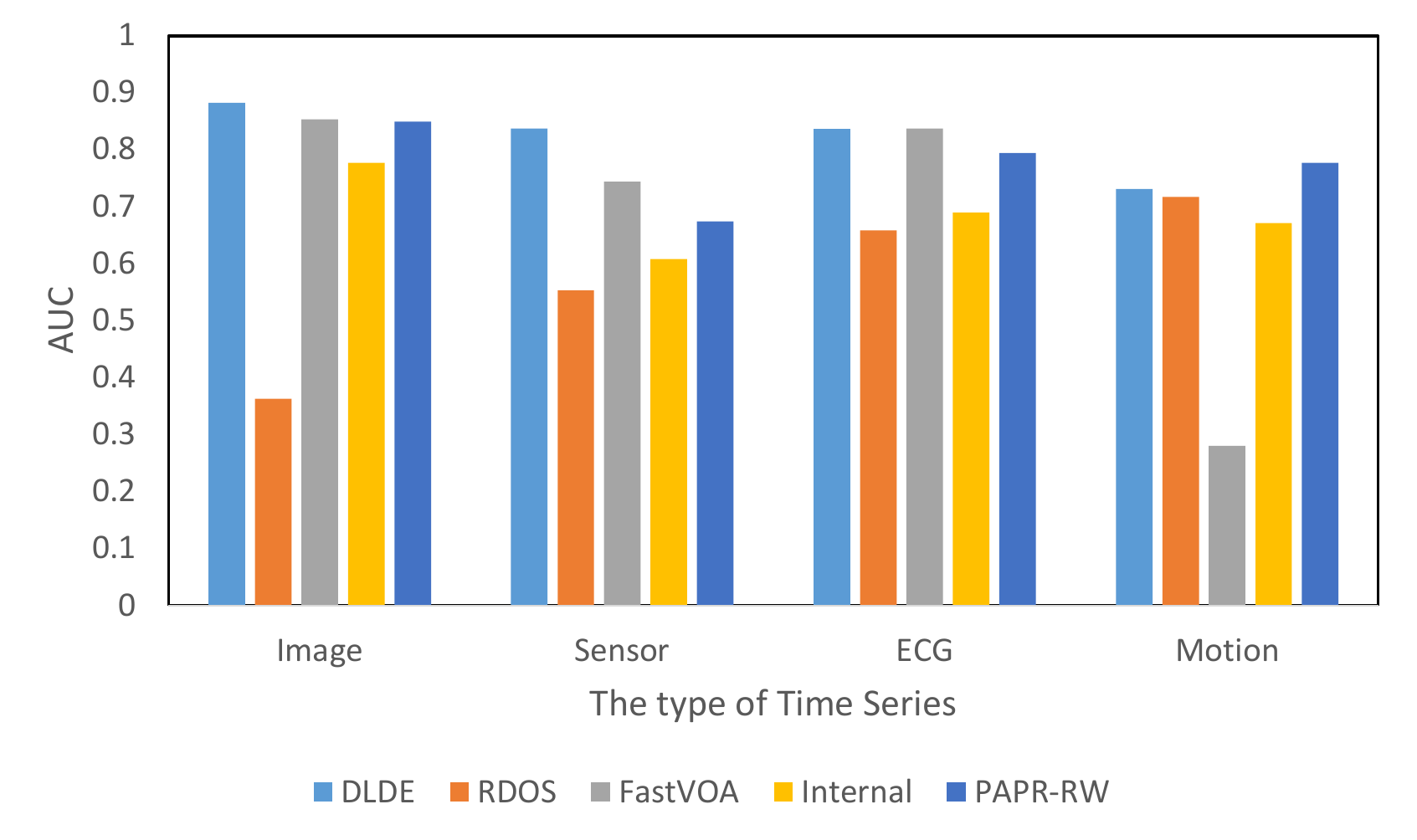}
	}
	\caption{Comparison of various experiments on AUC.}
	\label{fig:auc_type}
\end{figure}

For further analysis of experimental results, the data sets in Table \ref{table:ucr} are divided into four parts according to the length, the average of each four parts are the final results of each algorithm on different length of data sets.
The RDOS algorithm does not get running results on two data sets which are not be considered in the condition of more than 1000 part. From the experimental results in the Fig. \ref{fig:auc_type}(a), we can find that the algorithm DLDE can obtain better experimental results on time series data sets of different lengths. The results are shown in Fig. \ref{fig:auc_type}(b), it indicates that DLDE performs better on the first three types of data sets than other algorithms.


To test the impact of different data types on experimental results, the data set in Table \ref{table:ucr} is divided into four parts: ECG, MOTION, IMAGE, SENSOR.
ECG data and SENSOR data sets are typical time series data sets; MOTION is sequence data generated by actions; IMAGE data can extract time series data. The average of the experimental results of each algorithm on the four data sets is calculated separately as Fig. \ref{fig:auc_type}(b). It can be seen from the figure that DLDE performs better than the other algorithms on the first three types of data sets.


\subsection{Parameter analysis}
\begin{figure}[h]
	\centering
	\subfigure[Sensitivity analysis of DLDE to parameter $m$.]{
		\includegraphics[width=0.45\linewidth]{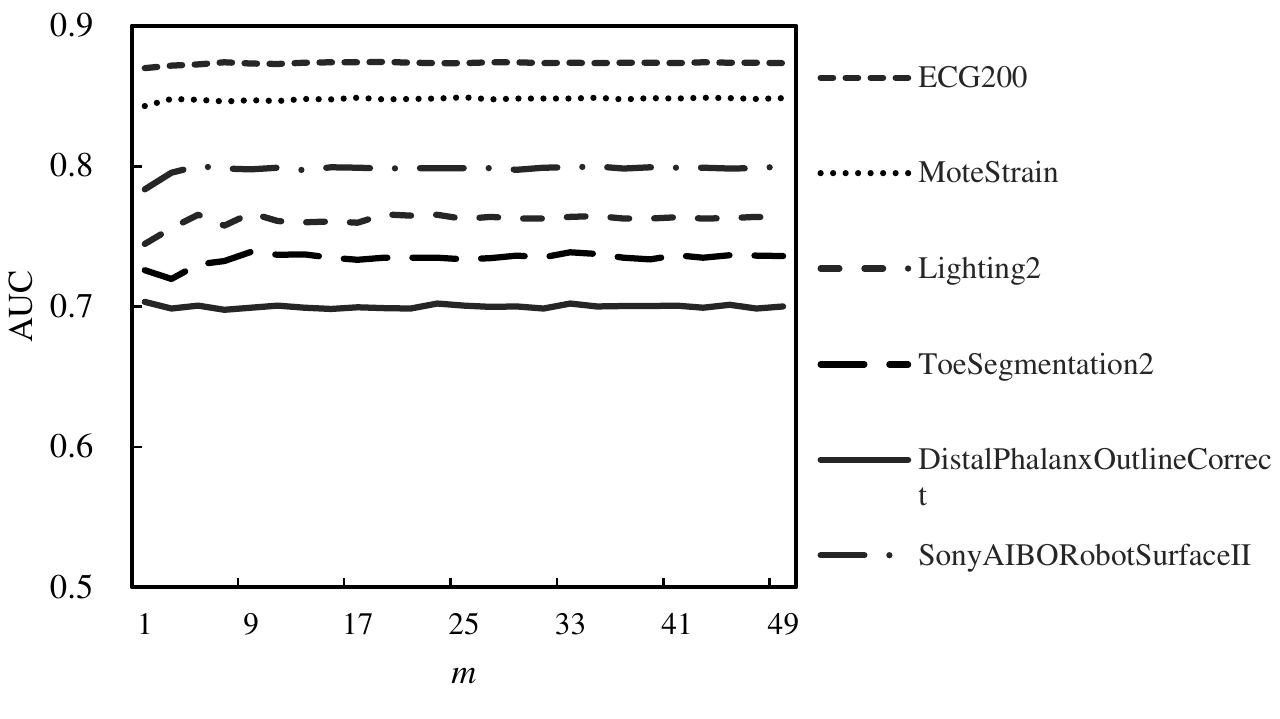}
	}
	\quad
	\subfigure[Sensitivity analysis of DLDE to parameter $h$.]{
		\includegraphics[width=0.45\linewidth]{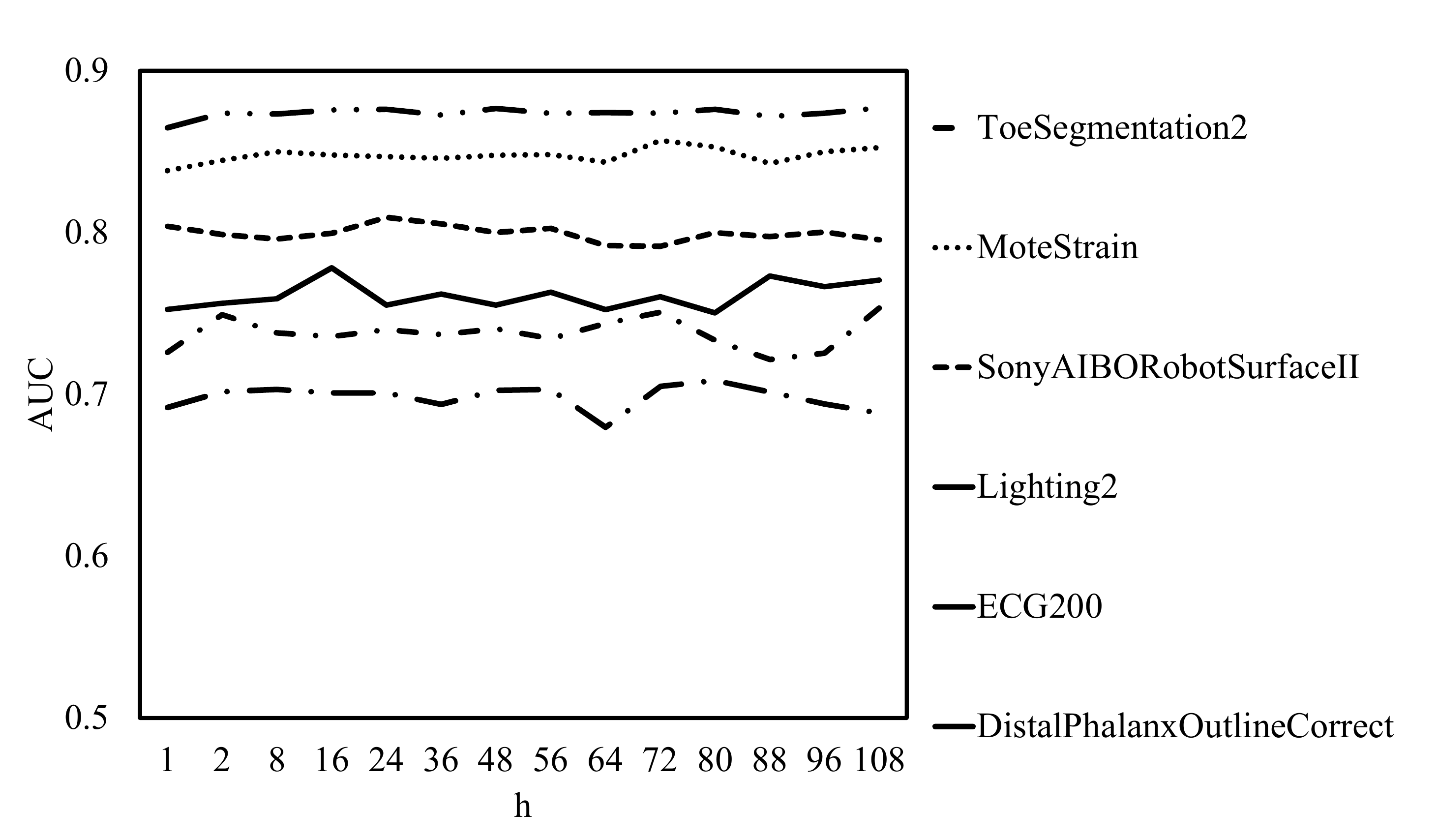}
	}
	\caption{The parameters analysis of dynamic window $m$ and hash number $h$.}
	\label{fig:mp}
\end{figure}

\textbf{Dynamic window $m$.}
In the DLDE algorithm, the dynamic time segment window is randomly divided, in order to ensure the stability of the algorithm, we choose to use the idea of ensemble learning. That is randomly divide $m$ times, and the final test result of the algorithm is the average of the number of runs. In this experiment, the sensitivity of the DLDE algorithm to the parameter $\textbf{m}$ will be verified. When $m$ is taken from 1 to 50, the variation of the AUC index on different data sets is recorded. Parameter m is tested under each parameter condition, the average value of the program running 50 times is taken as the final result and recorded in Fig. \ref{fig:mp}(a). In this figure, it can be noted that the experimental results of DLDE are basically in a stable trend on the data set of all algorithms, that is, the AUC index of the algorithm is basically convergent when m reaches 10. Therefore, this experiment proves that the algorithm does not require a lot of random division of dynamic time segments, and the algorithm has better stability.

\textbf{Hash number $h$.}
We construct a hash table at the leaf node with h p-stable local-sensitive hash functions\cite{7837870}. This hash defines a boundary region, and all values in this region have the same hash value. 
To avoid the instability of random hash functions, we use multiple random hash functions. The intersection of similar sets of data points computed by the generated plurality of hash functions is a final set of similar data points that can more accurately measure similar relationships between data points.
It is proved in the Fig. \ref{fig:mp}(b) that when $h$ is taken from 1 to 108, the algorithm can achieve convergence as long as m and h take a small constant.

%

\begin{figure}[tbh]
	\centering
	\includegraphics[width=0.65\linewidth]{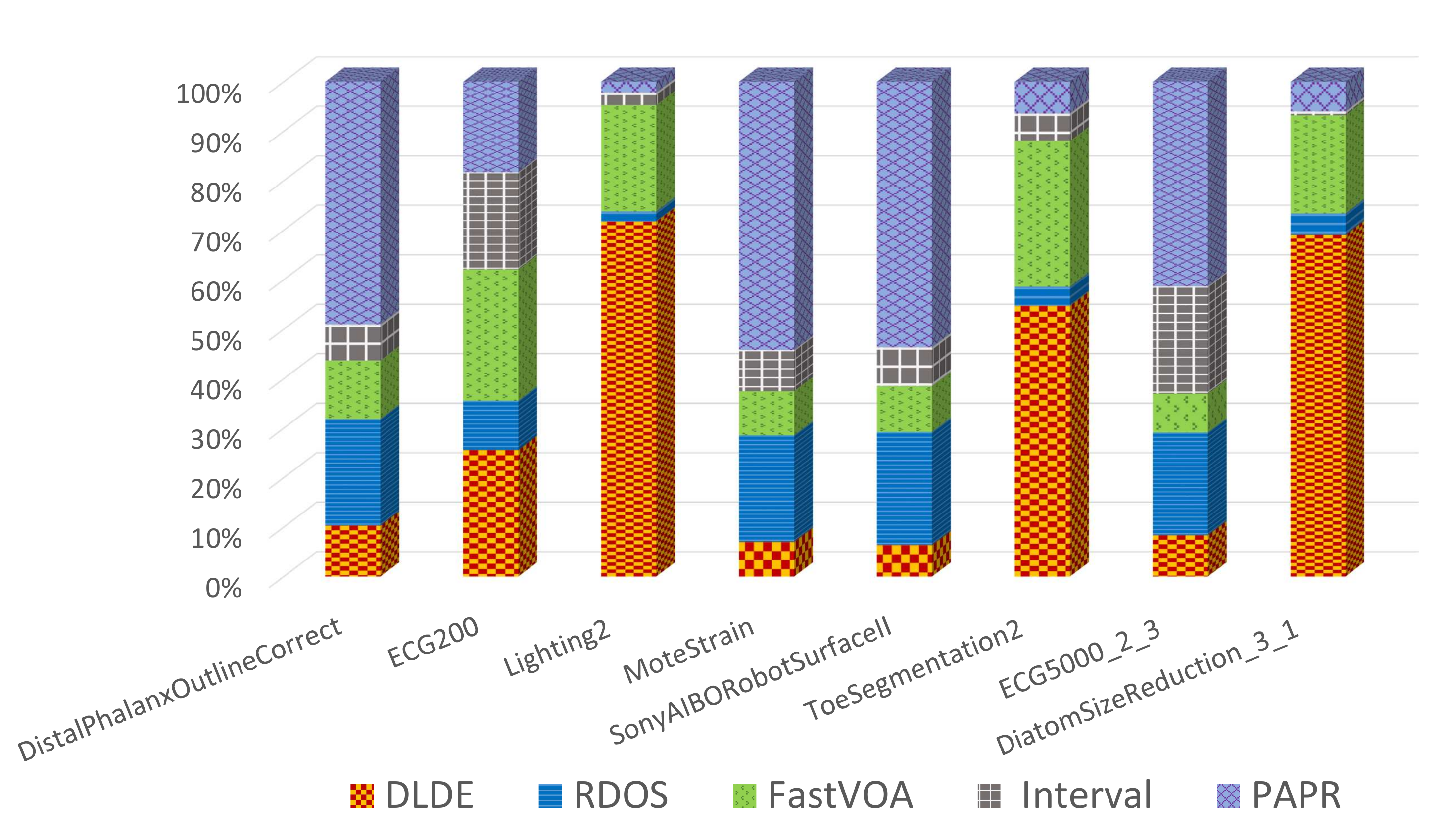}
	\caption{Comparison results of each algorithm on different length time series data}
	\label{fig:computation_time}
\end{figure}

\textbf{Computational time.}
In order to calculate the consumption time, we selected 8 data sets for testing. We calculate the percentage of calculation time for the five methods in each data set. As can be seen from the Fig. \ref{fig:computation_time}, DLDE has a better effect on the short length of the subsequence, and PAPR has a better effect on the long length of the subsequence, and the average performance of other data sets is relatively nearly.

\subsection{Performance on ECG data}	
In this section, we will demonstrate the effectiveness of our algorithm on ECG 
data selected from BIDMC Congestive Heart Failure Database. We select two ECG 
records from this database, $chfdb01\_275$ and $chfdb13\_45590$. 
These two ECG data contains two ECG signal, and each record contain one anomaly subsequence.

In this experiment, we use the data of one minute length in two data sets as experimental data, and divide the whole time series into 15 sub-time series according to the cycle per second. We will verify the difference between the scores calculated by the method proposed in this paper, and use the line graph to visualize this difference. Since the results calculated by our methods is the density of the subsequence, the score should convert into the abnormal score of the subsequence by using Equation (\ref{eq:anomaly}).
\begin{equation}
	anomaly\_score(P)=1-\sum\frac{p_i-min(P)}{max(P)-min(P)}
	\label{eq:anomaly}
\end{equation}

\begin{figure}[htbp]
	\centering
	\subfigure[ECG1 in chfdb01\_275]{
		\includegraphics[width=0.45\linewidth]{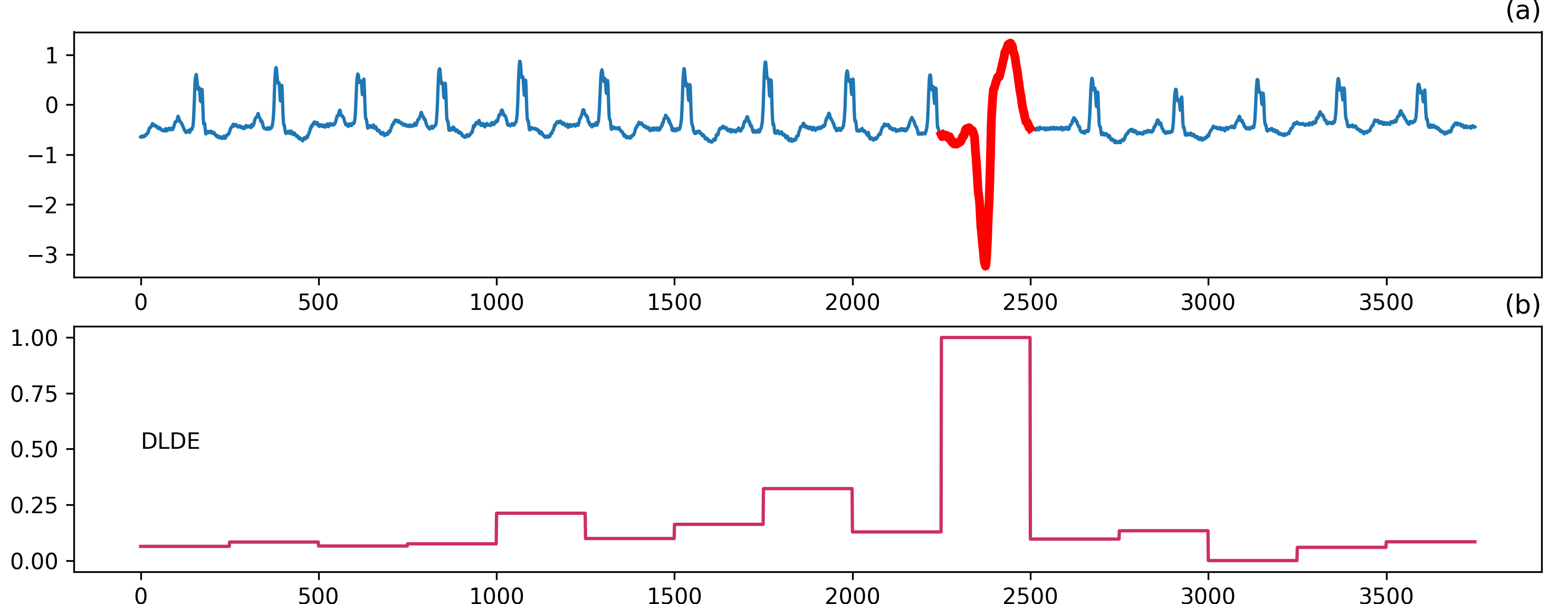}
		
	}
	\quad
	\subfigure[ECG2 in chfdb01\_275]{
		\includegraphics[width=0.45\linewidth]{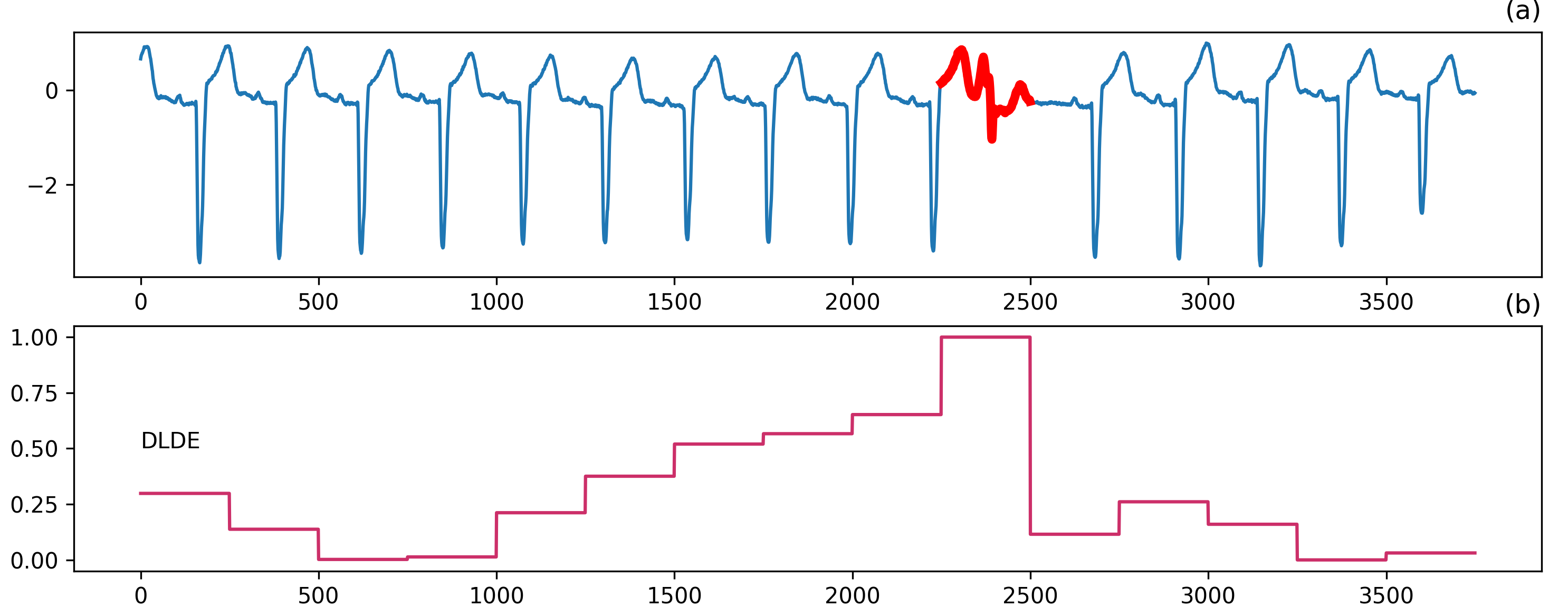}
	}
	\quad
	\subfigure[ECG1 in chfdb13\_45590]{
		\includegraphics[width=0.45\linewidth]{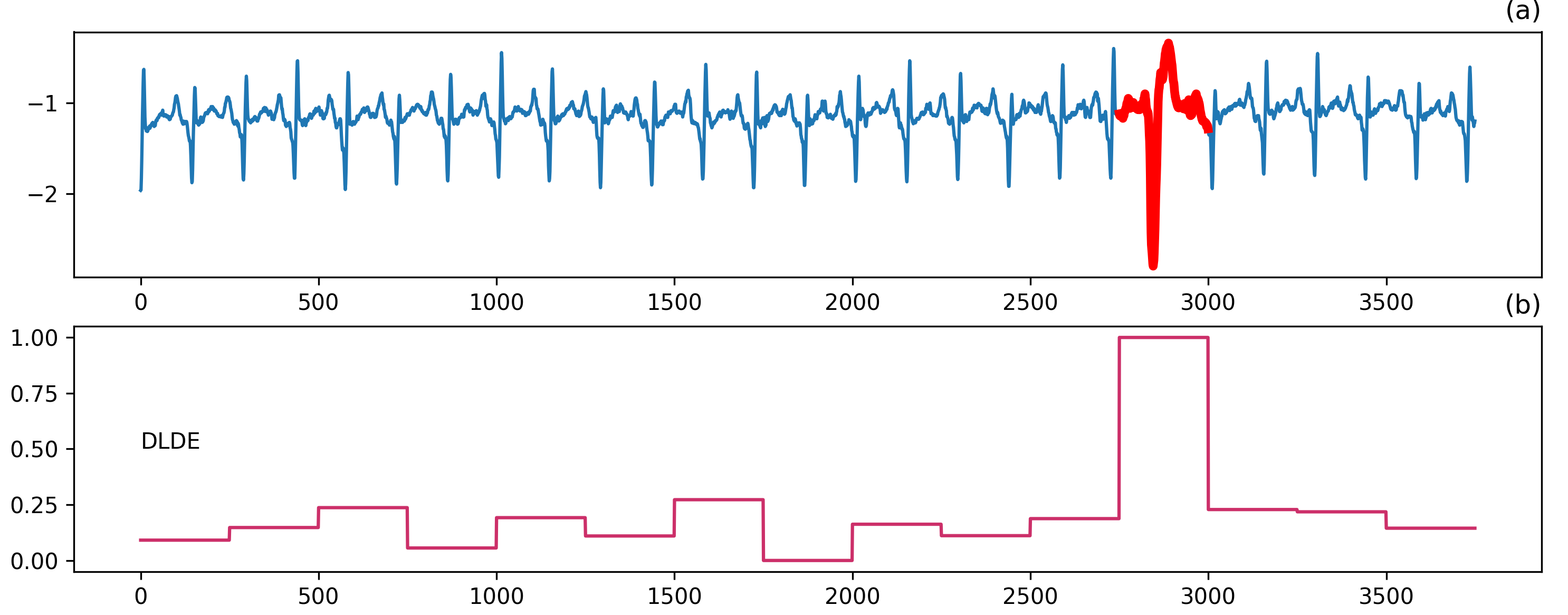}
	}
	\quad
	\subfigure[ECG2 in chfdb13\_45590]{
		\includegraphics[width=0.45\linewidth]{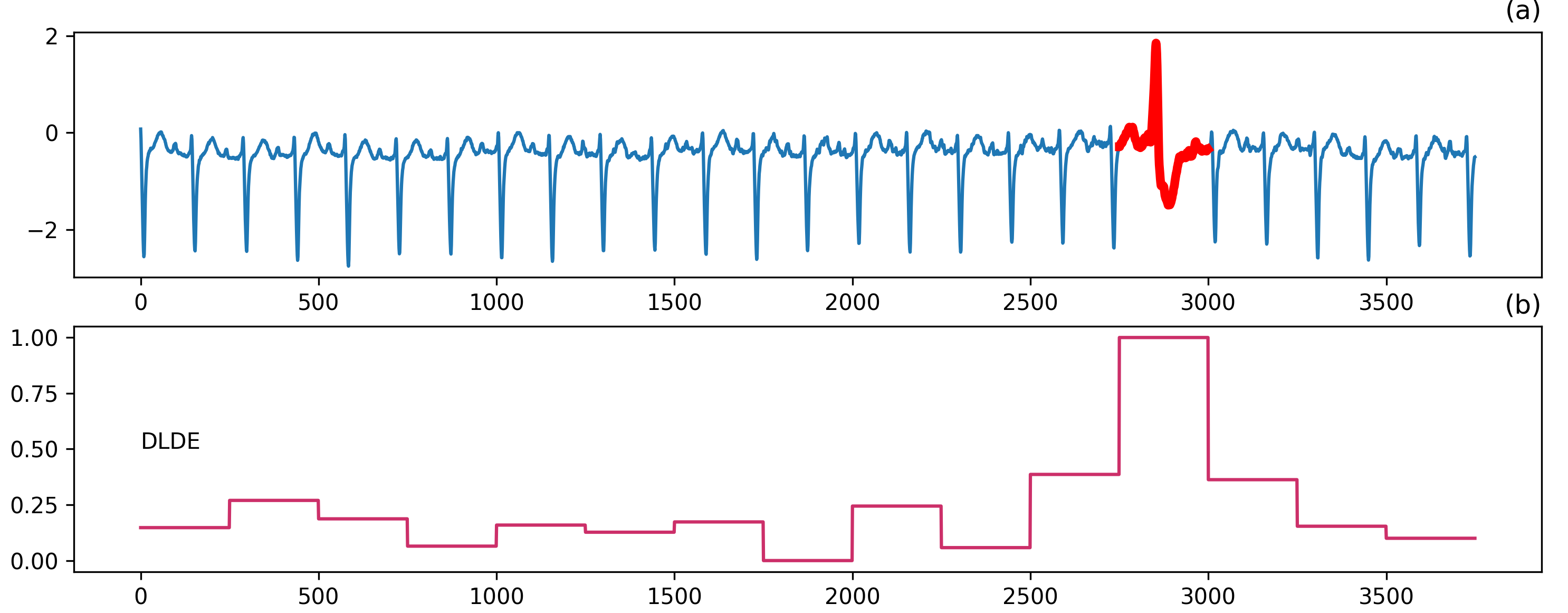}
	}
	\caption{The results on ECG data chfdb01\_275 and chfdb13\_45590}
	\label{fig:ecg275}
\end{figure}

These two ECG data are shown 
in Fig. \ref{fig:ecg275}, and the anomaly subsequences are shown in red 
line. The anomaly scores of each subsequence calculated by DLDE are shown in 
the dark red line below. It can be clearly found that the higher anomaly score is corresponded to true anomaly subsequence, and other score are around 0.5. We rank the anomaly scores of each subsequence to determine the anomalies in the data, thus avoiding the occurrence of missed detection. Therefore, these results can demonstrate the effective of our algorithm.

%

\section{Conclusion} \label{sec:conclusion}
In this paper, we propose a novel anomaly subsequence detection algorithm based on dynamic local density estimation(DLDE), which inspired by the idea of the similarity calculation method of dynamic time warping. The anomaly detection algorithm divides the time serieswith TSTree and uses the random hash function to quickly estimate the local density of the data points in the dynamic time segment. In order to avoid the randomness of dynamic time segments and hash functions, the idea of ensemble learning is adopted to ensure that the algorithm can obtain more stable detection results. 
Experimental results show that the proposed DLDE method performs better on different types of data sets than other baselines.
In the future work, we need to consider whether can set the automatic time segmentation method to reduce the process of algorithm ensemble learning.

\section*{Acknowledgment}
This study was supported by the Shenzhen Research Council(Grant No.
369 JSGG20170822160842949, JCYJ20170307151518535)

%
%
%
 \bibliographystyle{splncs04}
 \bibliography{ref}
%
%
%
%
%
\end{document}